\documentclass{article}
\usepackage[affil-it]{authblk}
\usepackage[legalpaper, margin=1.2in]{geometry}
\usepackage{graphicx}
\usepackage{times}
\usepackage{epsfig}
\usepackage{graphicx}
\usepackage{amsmath}
\usepackage{amssymb}
\usepackage{array}

\usepackage{xcolor}
\usepackage[normalem]{ulem}
\definecolor{dgreen}{rgb}{0,.6,0}

\newcolumntype{C}[1]{>{\centering\let\newline\\\arraybackslash\hspace{0pt}}m{#1}}

\begin{document}
\title{Stability of Internal States in Recurrent Neural Networks Trained on Regular Languages}
%
%
\author{Christian Oliva and Luis F. Lago-Fern\'{a}ndez}
%
%
\affil{Escuela Polit\'{e}cnica Superior, Universidad Aut\'{o}noma de Madrid\\
28049 Madrid, Spain\\
christian.oliva@estudiante.uam.es\\
luis.lago@uam.es}
%
\maketitle              
\begin{abstract}
We provide an empirical study of the stability of recurrent neural networks trained to recognize regular languages. When a small amount of noise is introduced into the activation function, the neurons in the recurrent layer tend to saturate in order to compensate the variability. In this saturated regime, analysis of the network activation shows a set of clusters that resemble discrete states in a finite state machine. We show that transitions between these states in response to input symbols are deterministic and stable. The networks display a stable behavior for arbitrarily long strings, and when random perturbations are applied to any of the states, they are able to recover and their evolution converges to the original clusters. This observation reinforces the interpretation of the networks as finite automata, with neurons or groups of neurons coding specific and meaningful input patterns.

\end{abstract}
\section{Introduction}
\label{sec::introduction}

The relationship between Recurrent Neural Networks (RNNs) and formal languages has been extensively studied since the early 90s \cite{DBLP:conf/nips/GilesMCSCL91,DBLP:journals/neco/ZengGS93}. It is well known that RNNs are able to learn different kinds of
regular and context independent languages \cite{casey_1998,gers_and_schmidhuber_2001,omlin_and_giles_1996}, and it has been demonstrated that they are equivalent to  Turing machines \cite{SIEGELMANN1995132}. Aiming at improving interpretability, many works in the literature have explored different methods to extract a Deterministic Finite Automaton (DFA), or a set of rules, from a RNN trained with positive and negative examples of a regular language \cite{casey_1998,cohen_2017,jacobsson_05,DBLP:journals/corr/abs-1709-10380}. However this approach has sometimes been criticized \cite{kolen}, since the process of mapping the network activation to a discrete set of states usually depends on an artificial quantization of a continuous space. Hence, in spite of the computational power of RNNs, it is still not clear whereas a RNN trained to recognize a formal language can be reduced to the corresponding abstract machine.

A more recent work \cite{DBLP:conf/icann/OlivaL19} has shown that the introduction of noise in the activation function of a RNN, together with proper regularization, forces the neurons to work in an almost binary regime. In that situation the activation of the recurrent units forms clusters that may be interpreted as states in a finite state machine \cite{DBLP:conf/icann/OlivaL19} without the need of an explicit quantization. Although the extracted automata seem to be valid, no in-depth analysis of the stability of the states was performed. 

In this article we build on their work adding the following contributions. First, we include a theoretical analysis of the effect of the noise on the discretization of the network activity. Second, we show that the networks are able to generalize to sequences that are several orders of magnitude longer than those used for training without any loss of accuracy. This is in contrast with the results obtained for traditional RNNs or LSTMs, for which a decrease of the accuracy is always observed as the string length increases. Finally, we analyze stability by introducing small perturbations that take the networks out of the activity clusters. We observe that the networks are able to assimilate these perturbations and move back to the original states recovering their initial behavior. Our results on several simple regular languages show that noise introduces stability in the networks, not only improving generalization but also contributing to the formation of stable activity clusters that are robust against small perturbations. These clusters can be mapped to discrete states in a DFA, showing that the networks learn to solve the problems by implementing abstract machines that are equivalent to the languages used for training. 

\section{Methodology}
\label{sec::methodology}


\subsection{Network description}
\label{sec::network}

Following the work of \cite{DBLP:conf/icann/OlivaL19}, here we use a simple Elman RNN \cite{elman_1990} with one single recurrent layer, $\tanh$ activation function and a noise term introduced into the preactivation:

\begin{equation}
\label{eq::recurrent_layer}
h_{t} = \tanh(W_{xh}x_{t} + W_{hh}h_{t-1} + h_{t-1} \circ X_{\nu} + b_{h}) \\
\end{equation}

\newpage

\noindent where $W_{xh}$ is the weight matrix connecting the input layer to the recurrent layer, $W_{hh}$ is the weight matrix in the recurrent connection, and $b_{h}$ is a bias vector. The noise term $h_{t-1} \circ X_{\nu}$ is an element-wise product between the activation in the previous time step $h_{t-1}$ and the vector $X_{\nu}$, whose components are randomly drawn from a normal distribution $N(0, \nu)$. The output $h_{t}$ of the recurrent layer is fed into an output layer with softmax activation function:

\begin{equation}
\label{eq::output_layer}
y_{t} = \sigma(W_{hy}h_{t} + b_{y})
\end{equation}

\noindent where $W_{hy}$ is the weight matrix and $b_{y}$ is the bias vector. 

\subsection{The effect of noise on the network activation}
\label{sec::noise}

\begin{figure}[t]
\begin{center}
   \hspace*{-0.5cm}
   \includegraphics[width=0.73\linewidth]{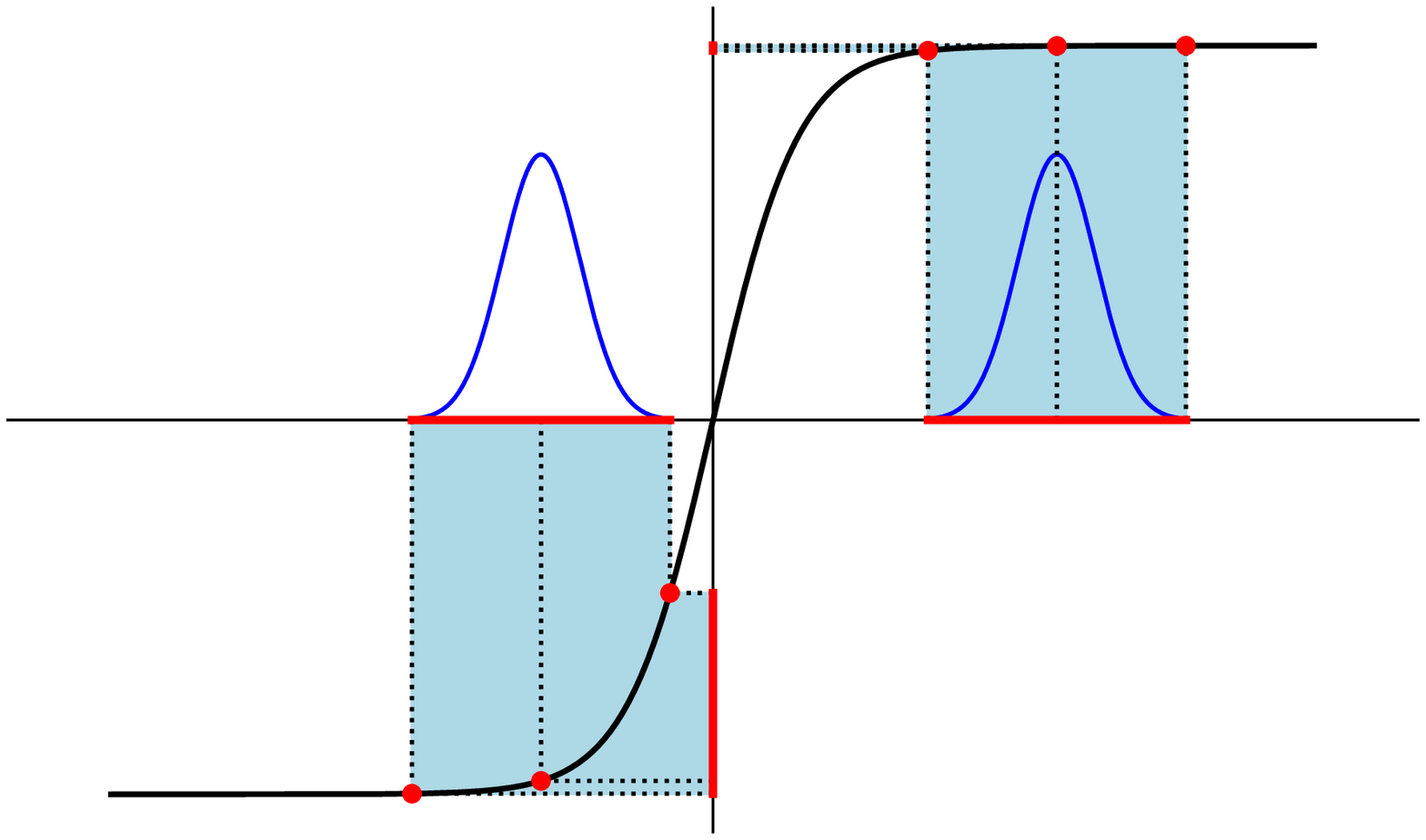}
   \includegraphics[width=0.23\linewidth]{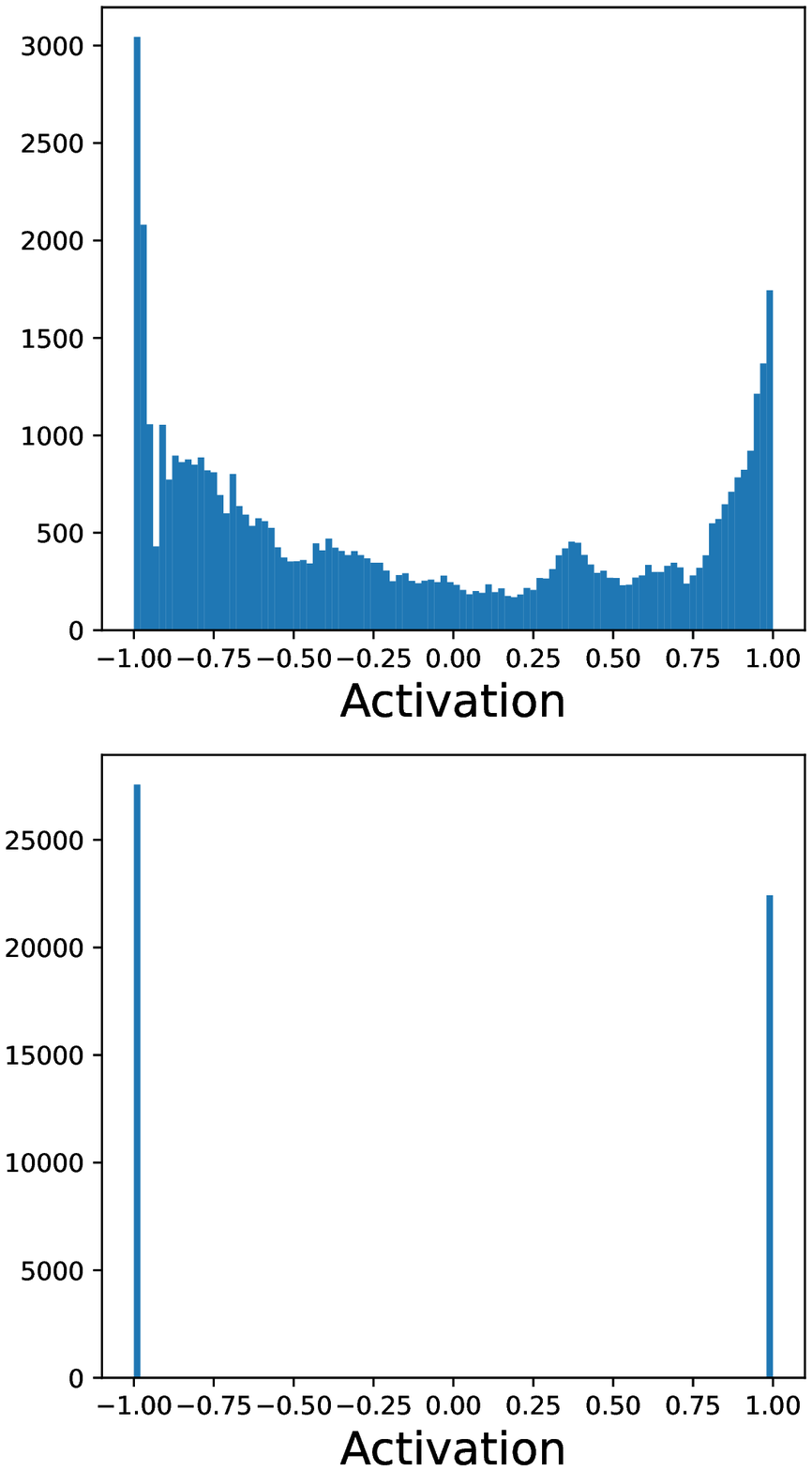}
\end{center}
   \caption{{\em Left.} The effect of pre-activation noise on the output of a $\tanh$ neuron. When the neuron is working close to the linear region, a small amount of noise in its input provides a very uncertain output. If the neuron is saturated, however, it can absorb the noise, providing a more stable output for the same amount of input noise. {\em Right.} Histograms of the activation of two typical neurons when no noise is used (top) and when noise with $\nu = 1.0$ is introduced into the pre-activation (bottom). Notice the difference in scale between the two histograms.}
\label{fig::noise_effect}
\end{figure}

The introduction of noise during neural network training has been frequently used as a regularizer \cite{doi:10.1162/neco.1995.7.1.108,analysis_of_noise_96}. It has also been reported that it may help the neurons escape from saturation, guaranteeing a better learning even with hard sigmoid or hard tanh activation functions \cite{DBLP:journals/corr/GulcehreMDB16}. Here we consider the possibility of using noise also to force the neurons into the saturation regime. As we show in figure \ref{fig::noise_effect} (left plot), if a neuron with $\tanh$ activation function needs to provide a stable response when noise is introduced into its pre-activation $z$, it needs to move towards the saturation region, where the noise is counterbalanced by the almost null slope of $\tanh(z)$. 

The noisiest region is of course that with the highest slope, so the neuron will try by all means to avoid staying close to $z=0$. If the networks are trained using regularization, however, there is a second element to take into account. In such a situation, the noise and the regularization have opposite effects: while the regularization tends to inactivate the neurons, pushing them towards zero activation, the noise pushes the neurons towards the saturation region. In order to avoid this tug of war, the noise enters equation \ref{eq::recurrent_layer} multiplied by the neuron's activity in the previous time step. This way a neuron that has been inactivated by the regularization is not affected by the noise. 

In the right side of figure \ref{fig::noise_effect} we show two typical histograms of a neuron's activation. The first case (top) corresponds to a neuron in a network trained without noise. The second one (bottom) is from a network trained with noise $\nu = 1.0$. It is clear from these histograms that the noise encourages neuron saturation: while the activation of the noiseless neuron spans the whole range $[-1, 1]$, the noisy neuron sits only at the extreme values. Neural networks trained with pre-activation noise will always display this behavior, with their neurons operating in an almost binary mode.

\subsection{Network training}
\label{subsec:networkTraining}

In all our experiments we use networks with $20$ units in the recurrent layer. The networks are trained to minimize a cross-entropy loss using a standard gradient descent optimizer with the Keras library \cite{lstm_keras}. L1 regularization with parameter $r = 0.0004$ is applied to all the network weights, but not to the biases. The networks are trained for $500$ epochs with a learning rate $l = 2.5$ and clipping the gradients to $c = 0.002$. The gradients are backpropagated through time for $s=25$ time steps. 

The noise parameter is set to $\nu = 1.0$ for most of the experiments. However for some tests we have observed that the use of a gradually increasing noise parameter, that starts being $0$ and linearly increases during training, helps to obtain better results. 

\subsection{Data}
\label{sec::data}

We consider two different kinds of problems that can be well modeled with finite automata: the recognition of regular languages and the addition of two integer numbers in numerical bases $2$ and $4$. 

\subsubsection{Regular languages.}

First, we consider several simple regular languages on the alphabet $\Gamma=\{a, b\}$ (see description in table \ref{tab::tomitaGramars}). They comprise the set of strings with an even number of appearances of the $a$ symbol, the set of strings starting with a $b$ symbol and ending with an $a$ symbol, and the seven \textit{Tomita Grammars} \cite{tomita}. The datasets for training and evaluating the networks are generated as described in \cite{DBLP:conf/icann/OlivaL19}, including the $\$$ symbol used as a separator. The input data are random strings built with the symbols $a$, $b$ and $\$$. For each input symbol the networks must predict if the partial substring starting at the last $\$$ symbol belongs or not to the language under consideration. Table \ref{tab::sum_example} shows an example of input-output strings for the Tomita 3 problem. A network's accuracy is evaluated by measuring the number of correct predictions on a validation set that is not used during the training phase. 

\begin{table}[t]
	\begin{center}
		\caption{Description of the $9$ regular languages used to train the networks.}
		\begin{tabular}{| l | l |}
			\hline
			\textbf{Problem} & \textbf{Regular language}  \\
			\hline
			parity & Strings with an even number of $a$'s. \\
			\hline
			bxa & Strings starting with $b$ and ending with $a$. \\
			\hline
			tomita1 & Strings with only $a$'s. \\
			\hline
		    tomita2 & Strings with only sequences of $ab$'s. \\
			\hline
			tomita3 & Strings with no odd number of consecutive $b$'s after an odd number \\
			        & of consecutive $a$'s. \\
			\hline
			tomita4 & Strings with fewer than 3 consecutive $b$'s. \\
			\hline
			tomita5 & Strings with even length with an even number of $a$'s. \\
			\hline
		    tomita6 & Strings where the difference between the number of $a$'s and $b$'s is a \\
			        & multiple of 3. \\
			\hline
			tomita7 & $b^{*}a^{*}b^{*}a^{*}$ \\
			\hline
		\end{tabular}
		\label{tab::tomitaGramars}
	\end{center}
\end{table}

\subsubsection{Addition in numerical bases $2$ and $4$.}

The second problem consists of predicting the result of a sum of two integer numbers that are presented to the network digit by digit. We consider additions in the numerical bases $2$ and $4$. At each time step the network is expected to return a single digit in the corresponding base which is the result (without carry) of the addition of the two inputs plus the possible carry generated in the previous time step. One example of input and output for base $4$ is shown in table \ref{tab::sum_example}, where the strings must be read from left to right, with the most significant digit at the rightmost position. Note that we also include the separator symbol $\$$ in the input strings, which represents the end of a particular sum. Whenever it appears, the $\$$ symbol is present in both input strings at the same time. The expected output for a $\$$ input is simply the last carry. As before, the input strings are randomly generated, and the networks are evaluated by measuring the number of correct predictions on a validation set. 

\begin{table}[t]
    \centering
    \caption{Example of input and output strings for the Tomita 3 and the base 4 addition problems.}
    \begin{tabular}{|l|l|}
        \multicolumn{2}{c}{Tomita 3}\\
        \hline
        Input   & \texttt{\$baa\$abaababbaabbaba\$bbbbbaabbb\$ab\$\$bab\$abbbbabbaababbbb} \\
        \hline
        Output  & \texttt{11111100000000000000111111111111101111011010110111110101} \\
        \hline
        \multicolumn{2}{c}{Base 4 addition}\\
        \hline
        Input 1 & \texttt{\$23111120100\$332001\$11010\$03333120\$23\$320021\$00321\$21223} \\
        \hline
        Input 2 & \texttt{\$31213111100\$212030\$03122\$22113023\$30\$100213\$22322\$30120} \\
        \hline
        Output  & \texttt{01103033120001111310102320211132001101030230122210112300} \\
        \hline
    \end{tabular}
    \label{tab::sum_example}
\end{table}

\subsection{Analysis}
\label{sec::analysis}

For all the above problems, we train the networks until they obtain a $100\%$ accuracy on both the training and validation datasets, which usually occurs in no more than $500$ training epochs. Once they are trained, we analyze the networks in terms of interpretability by inspecting the weight values and the activation space of the recurrent layer neurons. When we observe that the network activation forms well defined clusters, we analyze the stability of these clusters by: (1) measuring the network response to very long input strings that are several orders of magnitude longer than the strings used for training; and (2) introducing small perturbations into the activation clusters and observing if the network is able to recover its normal behavior.

\section{Results}
\label{sec::results}

In this section we describe mainly qualitative results that illustrate the behavior of RNNs trained with noise in the activation function. The networks are able to reach a $100\%$ accuracy in both the training and validation datasets for all the problems described above. Additionally, the activation in the recurrent layer is highly binarized in all cases, with the activation organized into a set of compact clusters and the networks admitting a clear interpretation as deterministic finite automata. Although this observation is general for all the considered problems, for illustration purposes we have selected the results for the Tomita 3 grammar and the base $4$ addition problem. 

\subsection{Activation in the recurrent layer}
\label{sec::results_activation_hidden_layer}

\begin{figure}[t]
\begin{center}
   \includegraphics[width=0.7\linewidth]{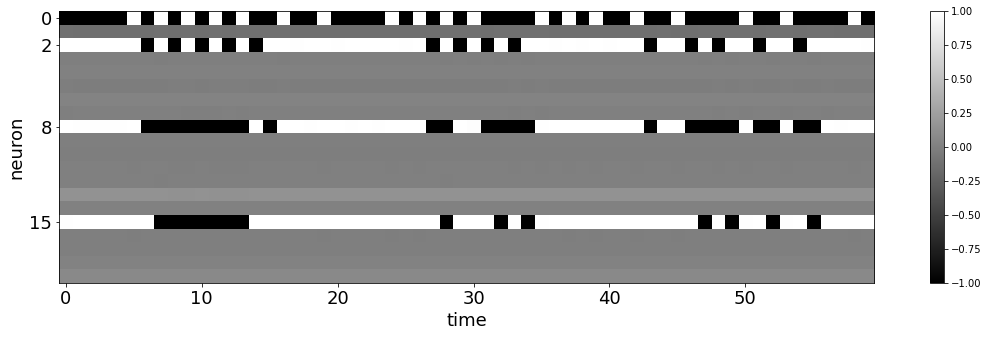} \includegraphics[width=0.26\linewidth]{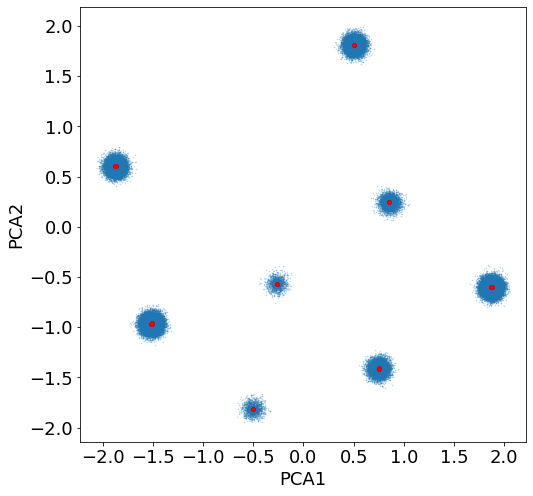}
   
   \includegraphics[width=0.7\linewidth]{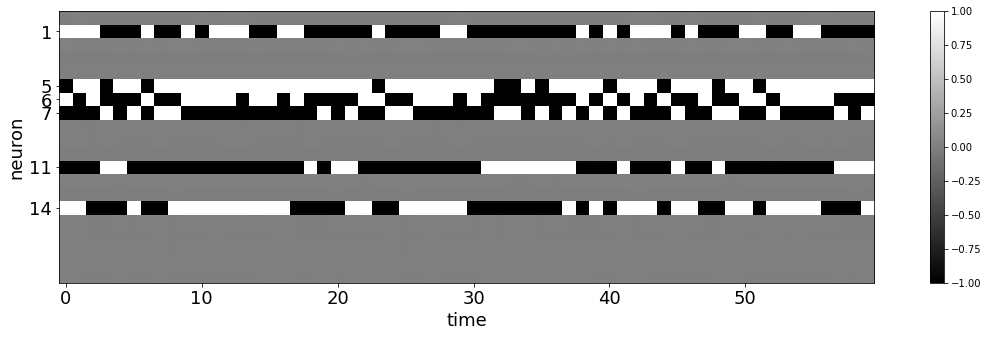} \includegraphics[width=0.26\linewidth]{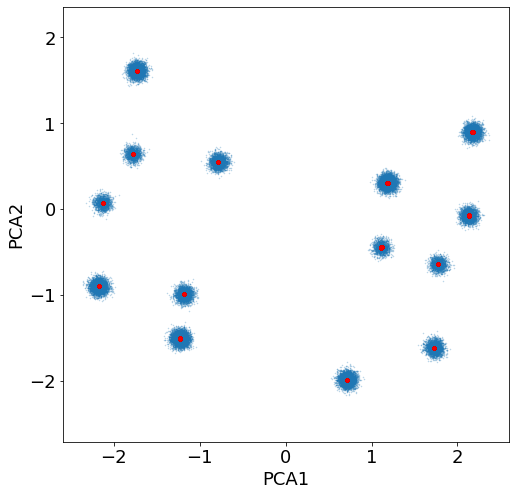}
\end{center}
   \caption{{\em Left.} Activation of the recurrent layer neurons for the first 60 input symbols in the test set in two networks trained on the Tomita 3 problem (top) and the base $4$ addition problem (bottom). In both cases only a small subset of the available neurons becomes activated, with the activation being always close to $+1$ or $-1$. The remaining neurons stay at a constant activation close to $0$. {\em Right.} Projection of the recurrent layer activation vector onto the first two principal components for the Tomita 3 problem (top) and the base $4$ addition problem (bottom). The activation vectors are clustered into $8$ and $14$ groups respectively. For the sake of a better visualization the clusters have been distorted adding a small Gaussian noise (blue dots), but the real clusters are much more compact (small red points inside the blue clusters).}
\label{fig::activation_with_pca}
\end{figure}

Figure \ref{fig::activation_with_pca} shows the activation in the recurrent layer for a network trained on the  Tomita 3 problem (top plots) and a network trained on the base 4 addition problem (bottom plots). On the left hand side we show the output of the $20$ recurrent neurons in response to the first $60$ symbols in the test set. Only a small fraction of these neurons are active, namely neurons $\{0, 2, 8, 15\}$ for the Tomita 3 problem and neurons $\{1, 5, 6, 7, 11, 14\}$ for the base 4 addition, displaying a completely binary output. The remaining neurons are always silent with an almost zero activation. As an additional way to visualize the recurrent layer activity, we have performed Principal Component Analysis of the activation vectors for the complete validation sets, which contain $100000$ input symbols for the Tomita 3 problem and $50000$ symbols for the addition. A scatter plot of the first two principal components is displayed on the right hand side of figure \ref{fig::activation_with_pca}, showing that the recurrent layer output self-organizes into a set of clusters, each corresponding to one of the unique binary activation vectors displayed as columns in the left side plots. There are $8$ clusters for the Tomita 3 problem and $14$ clusters for the base $4$ addition problem.

\subsection{Network weights}
\label{sec::network_weights}

\begin{figure}[t]
   \hspace*{0.5cm} \includegraphics[width=0.4\linewidth]{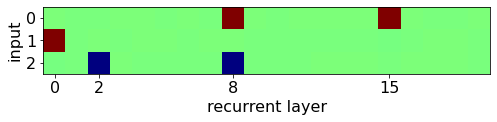} \hspace*{0.62cm} \includegraphics[width=0.4\linewidth]{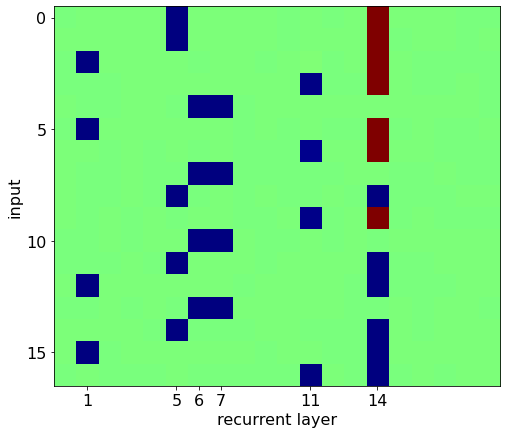} \\
   \hspace*{0.4cm} \includegraphics[width=0.41\linewidth]{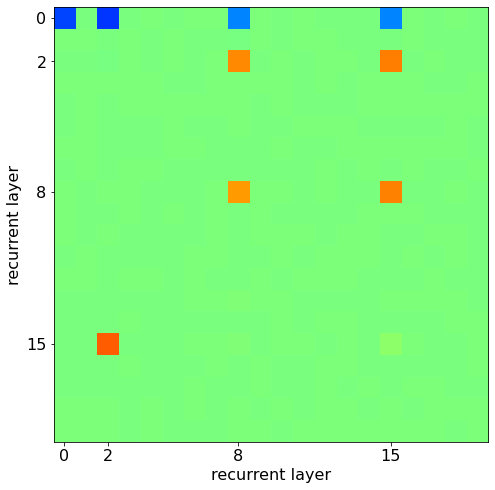} \hspace*{0.5cm} \includegraphics[width=0.48\linewidth]{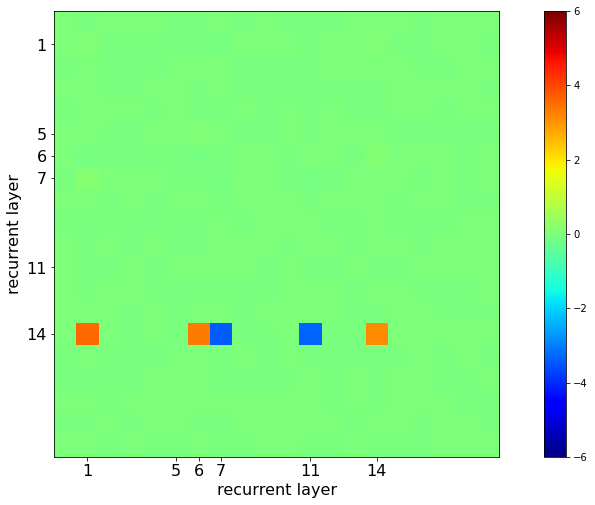} \\
   \caption{Color scale plots showing the network weights reaching the recurrent layer. {\em Top left.} Tomita 3, $W_{xh}$. {\em Bottom left.}  Tomita 3, $W_{hh}$. {\em Top right.} Base 4 addition, $W_{xh}$. {\em Bottom right.}  Base 4 addition, $W_{hh}$.}
\label{fig::network_weights}
\end{figure}

Weight visualization is also a good way of understanding the behavior of a recurrent neural network. In figure \ref{fig::network_weights} we plot in color scale the weight matrices $W_{xh}$ (input to recurrent layer, top) and $W_{hh}$ (recurrent layer to itself, bottom), for the Tomita $3$ problem (left) and the base $4$ addition problem (right). Only the weights involving the neurons that are active according to our previous analysis are significantly different from $0$. A very interesting observation regarding the addition problem is that only one recurrent neuron (neuron $14$) projects back onto the recurrent layer. This means that neuron $14$ is presumably coding the carry, which is the only information that must be kept from one time step to the next one in order to solve the addition problem.   

\subsection{Network interpretation as a finite state machine}
\label{sec::network_fsm}

\begin{figure}[t]
\begin{center}
   \includegraphics[width=0.9\linewidth]{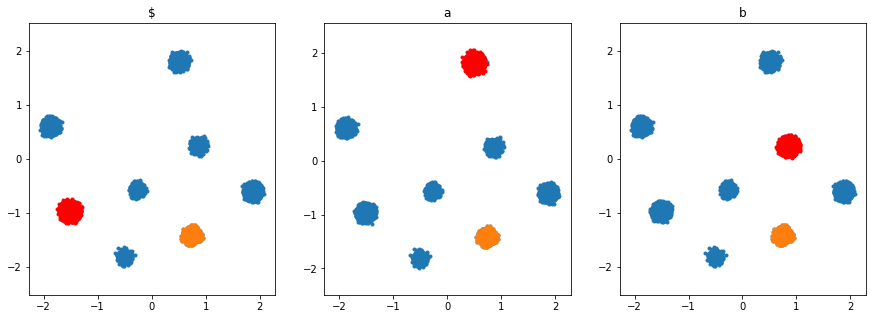}
\end{center}
   \caption{Example of deterministic transitions departing from an activation cluster in the network trained with the Tomita 3 data, in response to the input symbols $a$ (left), $b$ (middle) and $\$$ (right). Orange points represent the initial states, and they all belong to the same cluster. Red points are the destination states, which for a given input symbol also belong to the same cluster.}
\label{fig::tomita_3_transitions}
\end{figure}

Analyzing the transitions from each of the activation clusters in response to each of the input symbols, we observe that these transitions are deterministic. That is, if we initialize the recurrent layer to any of the points of one of the clusters shown in figure \ref{fig::activation_with_pca} and we observe the recurrent layer output after the network has processed one input symbol, we see that all the outputs belong again to the same cluster. Hence, in response to a given input sequence the network jumps from cluster to cluster just as a deterministic finite automata. Figure \ref{fig::tomita_3_transitions} shows an example of the transitions departing from one of the activation clusters in the Tomita 3 problem with the three input symbols $a$, $b$ and $\$$. Note that, as before, a small Gaussian noise has been added to all the points in order to improve the visualization. The real clusters are more compact than shown. 

\begin{figure}[t]%
  \centering
\begin{tabular}{|c|c|c|c|}
\hline
  & a & b & \$ \\
\hline\hline
0* & 7 & 5 & 3 \\
1 & 6 & 1 & 3 \\
2* & 7 & 2 & 2 \\
3* & 7 & 4 & 2 \\
4* & 7 & 4 & 2 \\
5 & 6 & 0 & 2 \\
6 & 1 & 1 & 3 \\
7* & 3 & 5 & 3 \\
\hline
\end{tabular}
%
  \hspace*{0.2cm} 
\begin{tabular}{|rlr|c|c|c|}
\hline
& & & a & b & \$ \\
\hline\hline
& 0 & (1, 6) & 0 & 0 & 4 \\
& 1 & (5) & 0 & 2 & 4 \\
& 2* & (0) & 3 & 1 & 4 \\
& 3* & (7) & 4 & 1 & 4 \\
$\rightarrow$ & 4* & (2, 3, 4) & 3 & 4 & 4 \\
\hline
\end{tabular}
  $\vcenter{\hbox{\includegraphics[width=0.48\linewidth]{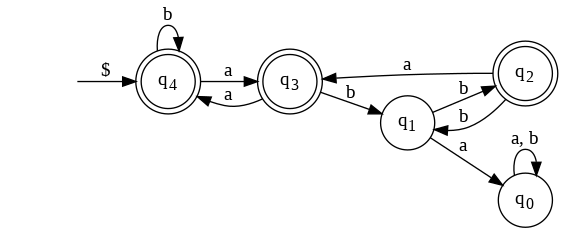}}}$
  \caption{{\em Left.} Transition table. States marked with an asterisk are acceptance states. {\em Middle.} Transition table after minimization. States marked with an asterisk are final states. The state marked with an arrow is the initial state. {\em Right.} Minimum DFA.}%
  \label{fig::table}%
\end{figure}

\paragraph{Automata extraction.} The activation clusters in the recurrent layer may be interpreted as the states of a deterministic finite automaton. Then it is not difficult to use transition plots like the one shown in figure \ref{fig::tomita_3_transitions} to obtain the transition table and, from this table, extract the corresponding DFA. The left panel in figure \ref{fig::table} shows the table obtained for the network trained with the Tomita $3$ data, displaying the transitions from the $8$ clusters/states with each of the $3$ input symbols. It is not difficult to see that some of these states are equivalent. If we apply a DFA minimization algorithm \cite{10.5555/1196416} we obtain an automaton with just $5$ states whose transition table is shown in the middle panel of figure \ref{fig::table}. The right panel of this figure also shows the transitions graph of the minimum DFA. 

The extracted automaton is in fact the minimum DFA for the Tomita 3 language, and this observation is repeated for all the problems that we considered. To summarize, in all cases we were able to reduce the trained networks to a DFA that recognizes the regular language that was used as input. The DFA states appear automatically when the networks are trained with noise in the activation of the recurrent neurons, and are not a result of artificial quantization of the hidden layer activation space. 

\section{Stability}
\label{sec::stability}

Although the results presented in section \ref{sec::results} are very interesting from the point of view of interpretability, the efforts made for reducing a recurrent neural network to a set of rules have been frequently criticized \cite{kolen}. One major problem here is the stability of the extracted states. If the states are not stable, a small deviation of the network activation with respect to the expected state may cause divergence in the long term, with unexpected behavior for very long input strings or inputs that differ substantially from the training samples.
In this section we perform a study of the stability of the trained networks, by analyzing their response to very long inputs and studying their behavior when small perturbations are introduced. 

\subsection{Response to very long strings}
\label{sec::long_strings}

\begin{figure}[t]
\begin{center}
   \includegraphics[width=0.95\linewidth]{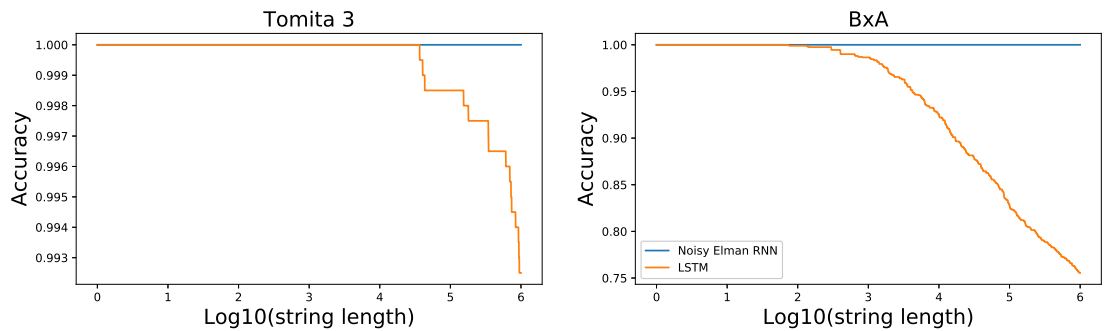}
\end{center}
   \caption{Accuracy versus string length for the noisy RNN ($\nu = 1.0$) and a standard LSTM network on the Tomita 3 problem (left) and the BxA problem (right). Note the logarithmic scale in the horizontal axis.}
\label{fig::unmillon}
\end{figure}

For this experiment we use additional test data sets with very long input strings ($10^{6}$ characters) where the $\$$ symbol appears only once, in the first position. Note that the typical training and test strings are in no case longer than $100$ characters, this means that we are testing the networks with strings that are $4$ orders of magnitude longer than those used for training. For each problem, we train $20$ different networks, which in all cases get $100\%$ accuracy on the standard validation sets. Then we test these networks on $100$ different {\em long} strings, and analyze their average accuracy for each input symbol. Figure \ref{fig::unmillon} shows a plot of the network accuracy versus the position of the symbol in the input string for the Tomita 3 and the BxA problems. The latter problem has been included in this section because it requires that the network remembers the first input symbol for all later time steps, and so it is more challenging than the other problems. To put these results in a proper context, the solution obtained with a standard LSTM network \cite{lstm} is also included. 

The most remarkable observation is that the noisy RNN networks are able to maintain the $100\%$ accuracy regardless of the string length. In contrast, the LSTM starts failing for string lengths around $3 \cdot 10^{5}$ for the Tomita 3 problem, and much earlier for the BxA problem. This result suggests that the activation clusters observed in the recurrent layer are in fact stable and that the DFA interpretation may be valid. 

\subsection{State perturbation}
\label{sec::state_perturbation}

\begin{figure}[t]
\begin{center}
   \includegraphics[width=0.9\linewidth]{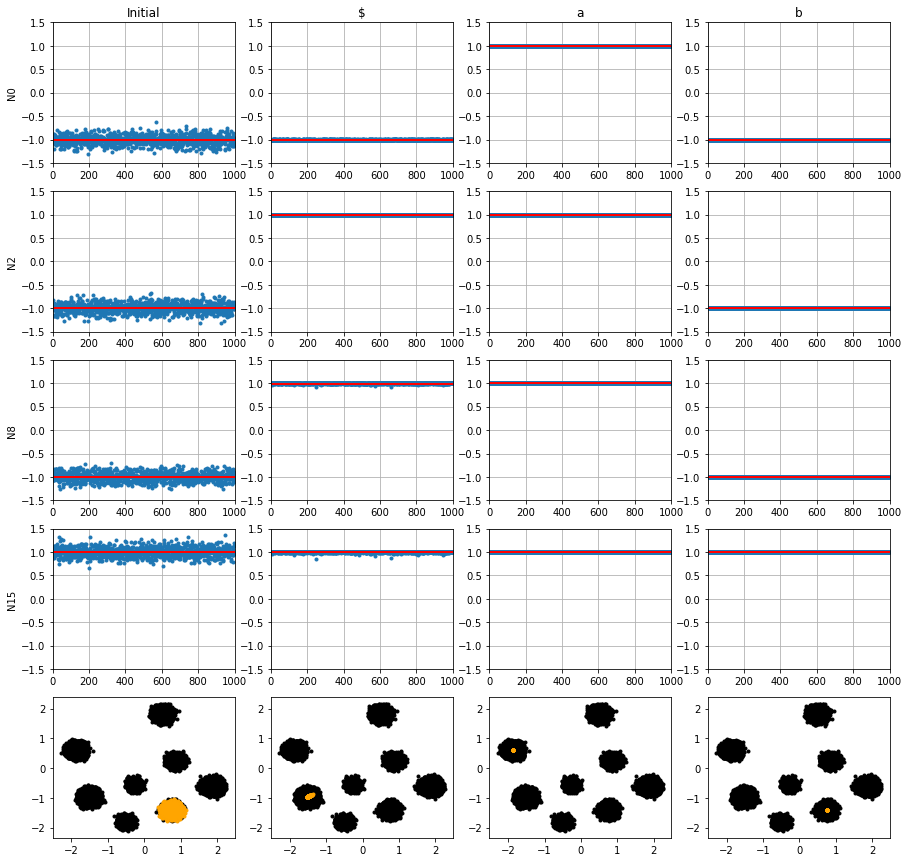}
\end{center}
   \caption{Activation in the recurrent layer of a network trained with the Tomita $3$ data. The first $4$ rows display the output of each of the $4$ active neurons, in the initial state and after processing the input symbols $\$$, $a$ and $b$, for $1000$ random tests. For each test, the recurrent neurons are first placed in a specific state (red lines in the first column) and then perturbed with random noise. Red lines in the plots represent the expected outputs, blue dots are the actual activations. The last row displays the recurrent layer activation (orange) superimposed to the activation clusters (black).}
\label{fig::tomita_3_transitions_sequence}
\end{figure}

After the analysis presented in the previous section, it seems that the activation clusters observed in the recurrent layer are indeed stable. In this section we test this hypothesis analyzing the network behavior when small perturbations are introduced into the clusters. With this purpose, we initialize the networks in one of the observed clusters, introduce some Gaussian noise and measure the activation in the recurrent layer as the input symbols are presented. Figure \ref{fig::tomita_3_transitions_sequence} shows one example for the Tomita $3$ problem. Each of the plots in the first $4$ rows shows the output of one of the $4$ neurons that are active in the recurrent layer in $1000$ different tests. The red lines represent the expected activation while the blue dots are the actual values. Each column represents one different time step, with the first one being the initial situation and the others showing the activation after presentation of the input symbols $\$$, $a$ and $b$. The figure shows that, in spite of the noise introduced in the initial states, the network is able to converge to the expected solution in just a few time steps. Note that the network can recover from the perturbation even though some neurons are forced to be initially out of the valid range $[-1, 1]$.

The last row in figure \ref{fig::tomita_3_transitions_sequence} shows the recurrent layer activation (orange) superimposed to the activation clusters (black). Due to the initial perturbation, the activation in the first column presents a high dispersion, but in just two time steps it converges to the center of the corresponding cluster. It is important to note that, regardless of the initial cluster and the input symbol, the activation always converges to center of another cluster (see figures \ref{fig::tomita_3_all_transitions}, \ref{fig::suma_4_all_transitions_a} and \ref{fig::suma_4_all_transitions_b} in the appendix). This means that, as far as the network is initialized to start inside one of the these clusters, it will continue jumping from one cluster to another forever. That is, the activation clusters observed in the recurrent layer are effectively working as the states in an equivalent DFA. 

\begin{figure}[t]
\begin{center}
   \includegraphics[width=0.45\linewidth]{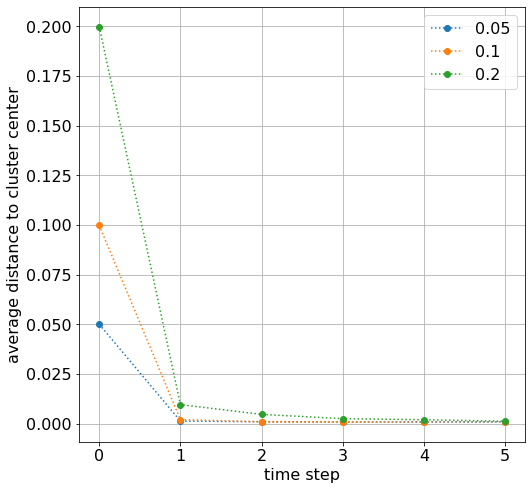}
   \includegraphics[width=0.45\linewidth]{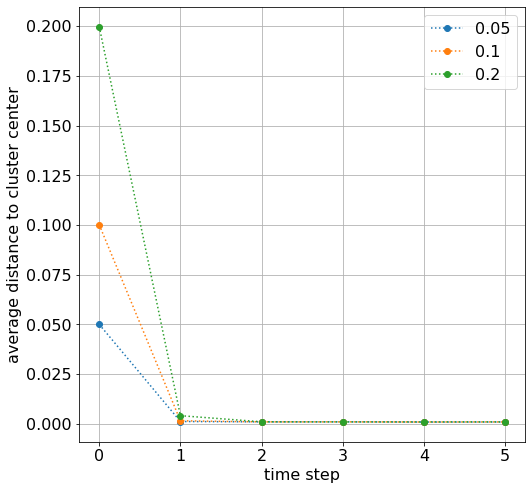}
\end{center}
   \caption{Average dispersion of the recurrent layer activation vectors with respect to the expected cluster centers versus time, for networks trained with the Tomita $3$ data (left) and the base 4 addition data (right). Three different perturbation levels ($0.05$, $0.1$ and $0.2$) are considered. 
   }
\label{fig::tomita_3_dispersion}
\end{figure}

In order to quantify this observation, we finally show in figure \ref{fig::tomita_3_dispersion} the average dispersion of the activation vectors with respect to the cluster centers for $5$ time steps after initialization and different perturbation levels. All possible initial clusters and all input strings of length $5$ have been considered. Although the initial dispersion is quite high, in just a few iterations the dispersion reduces to almost zero, indicating the convergence of the activation vectors to the cluster centers and reinforcing the idea of stability.

\newpage
\section{Conclusions}
\label{sec::conclusions}

In this article we have explored the use of noise in a recurrent neural network to force the neurons into the saturation region of their activation function, where their response is more binary and interpretable. When these noisy RNNs are trained with regular languages, the activation in the recurrent layer becomes a discrete set of clusters or states and, when an input sequence is presented to the network, its time evolution is reduced to a transition between states in response to each input symbol. This behavior resembles the operation of a finite state machine.

The stability of the activation clusters has been tested in two different ways. First, we have shown that the networks are able to generalize to sequences several orders of magnitude longer than those used for training with no apparent accuracy reduction. Second, we have studied the resilience of the networks when disturbing the activation clusters with small random perturbations, and we have shown that they are always able to assimilate these perturbations and recover the initial behavior.

In essence, networks trained with noise in the activation function self-organize into a set of well defined, discrete and stable activity clusters which are robust against small perturbations. They obtain a perfect generalization and admit a direct interpretation as deterministic finite automata. 

%
%
%
\bibliographystyle{splncs04}
\bibliography{ecml2020}
%
\appendix
\newpage
\newpage
\section{State transitions}

In this appendix we include additional figures showing the transitions starting from all the clusters/states in the Tomita 3 problem and the base 4 addition problem.


\begin{figure}[!bht]
\begin{center}
   \includegraphics[width=0.45\linewidth]{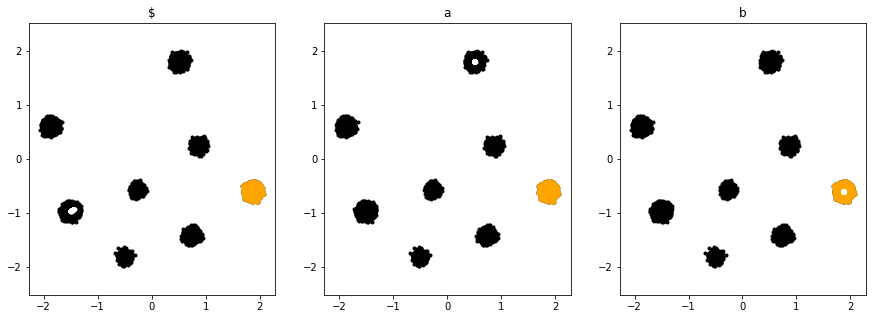} \hspace*{0.05\linewidth}
   \includegraphics[width=0.45\linewidth]{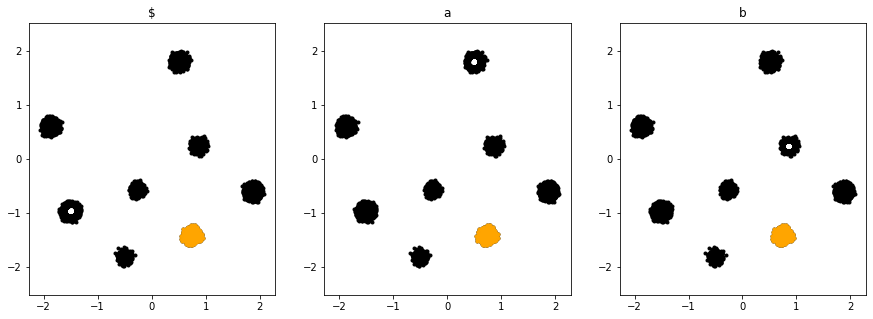}
   \includegraphics[width=0.45\linewidth]{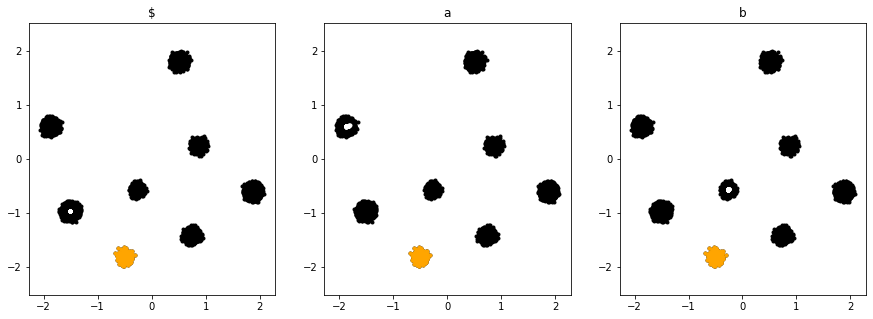} \hspace*{0.05\linewidth}
   \includegraphics[width=0.45\linewidth]{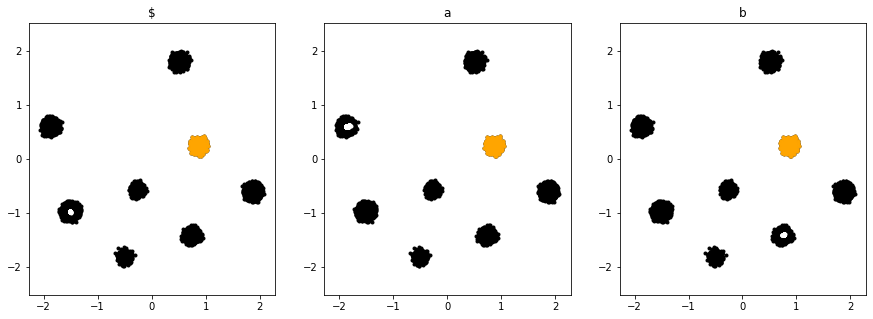}
   \includegraphics[width=0.45\linewidth]{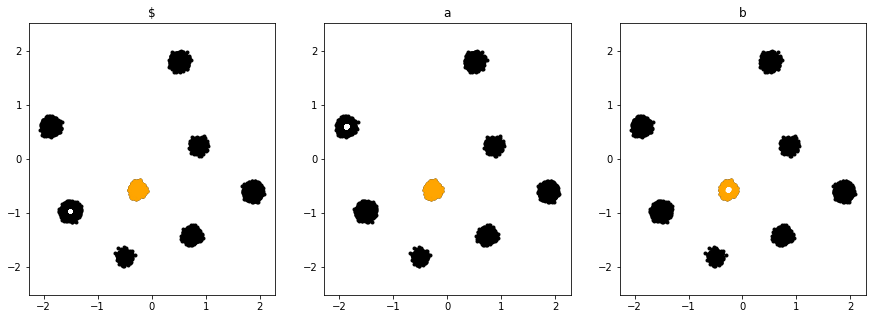} \hspace*{0.05\linewidth}
   \includegraphics[width=0.45\linewidth]{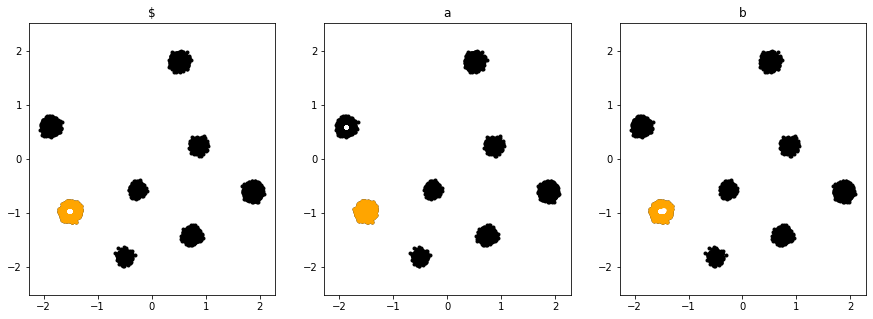}
   \includegraphics[width=0.45\linewidth]{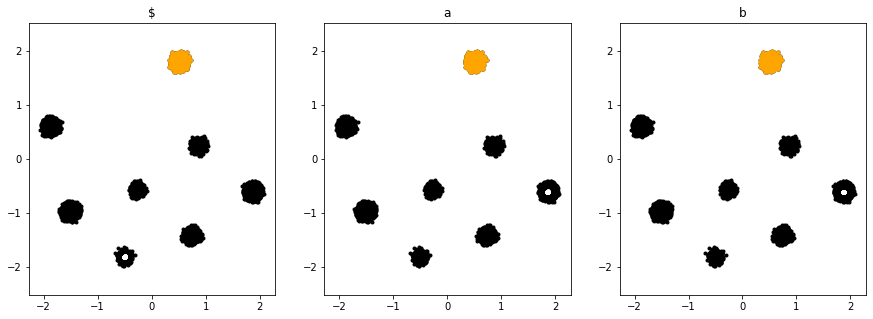} \hspace*{0.05\linewidth}
   \includegraphics[width=0.45\linewidth]{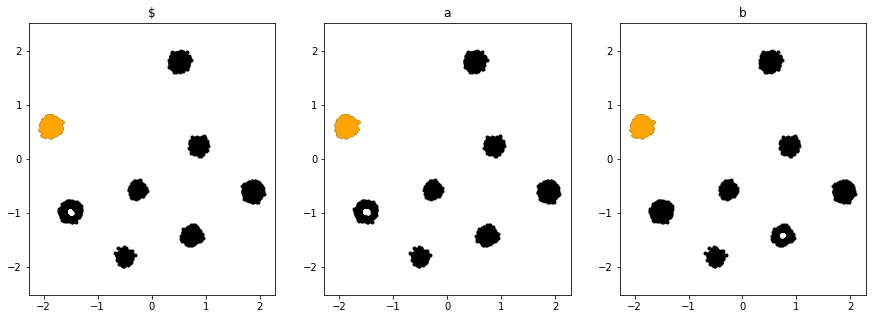}
\end{center}
   \caption{State transitions for the network trained with the Tomita $3$ data. The orange dots represent the starting points, the white dots represent the destinations. Each group of three plots represents the transitions from a given cluster/state with each of the input symbols ($\$$, $a$, $b$).}
\label{fig::tomita_3_all_transitions}
\end{figure}


\begin{figure}[!bht]
\begin{center}
   \includegraphics[width=0.8\linewidth]{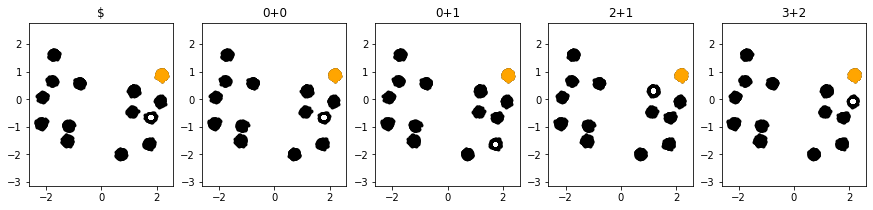} 
   \includegraphics[width=0.8\linewidth]{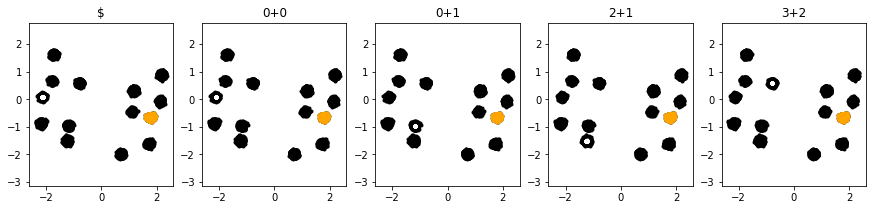}
   \includegraphics[width=0.8\linewidth]{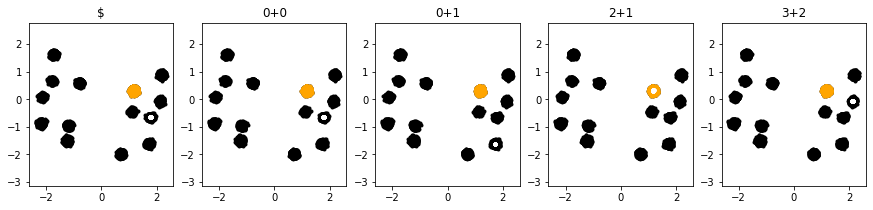} 
   \includegraphics[width=0.8\linewidth]{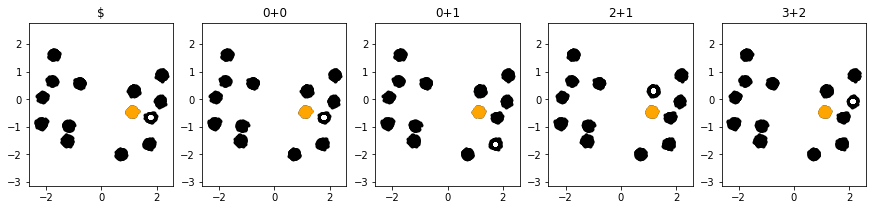}
   \includegraphics[width=0.8\linewidth]{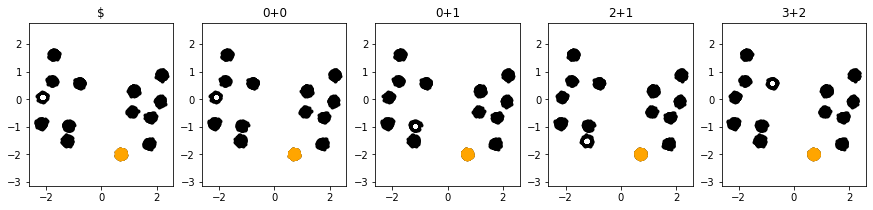}
   \includegraphics[width=0.8\linewidth]{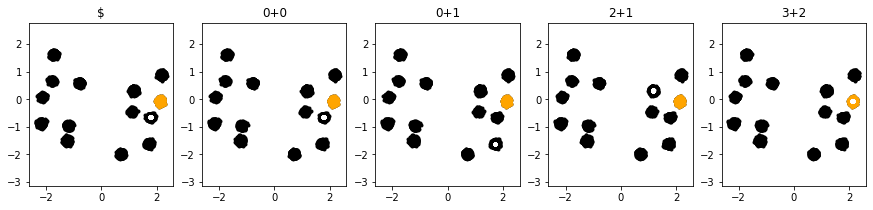}
   \includegraphics[width=0.8\linewidth]{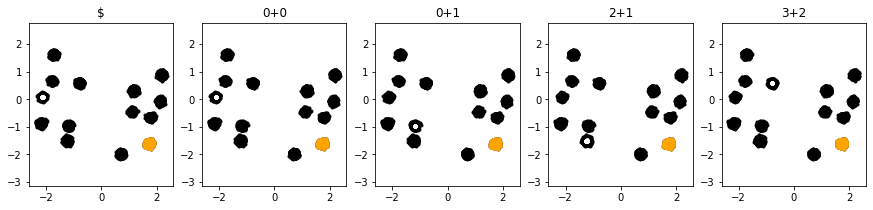} 
\end{center}
   \caption{State transitions for the network trained with the base $4$ addition data. The orange dots represent the starting points, the white dots represent the destinations. Each row represents the transitions from one of the first $7$ clusters/states. Each column represents the transitions with an input symbol. Only $5$ different symbols are shown.}
\label{fig::suma_4_all_transitions_a}
\end{figure}

\begin{figure}[!bht]
\begin{center}
   \includegraphics[width=0.8\linewidth]{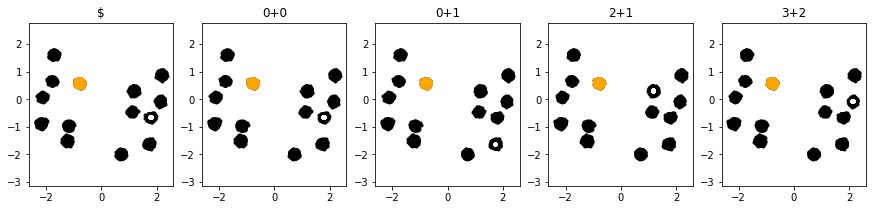} 
   \includegraphics[width=0.8\linewidth]{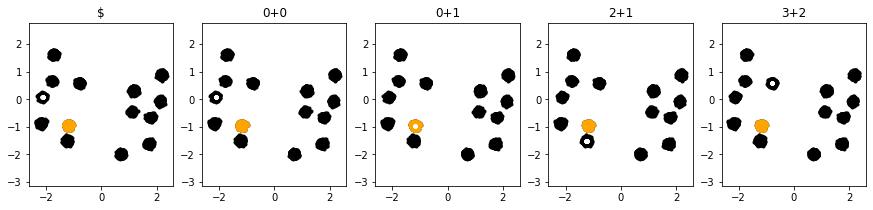}
   \includegraphics[width=0.8\linewidth]{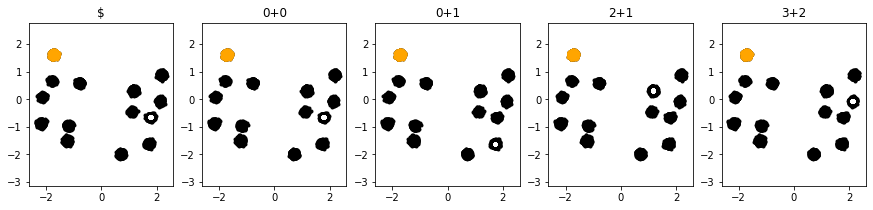} 
   \includegraphics[width=0.8\linewidth]{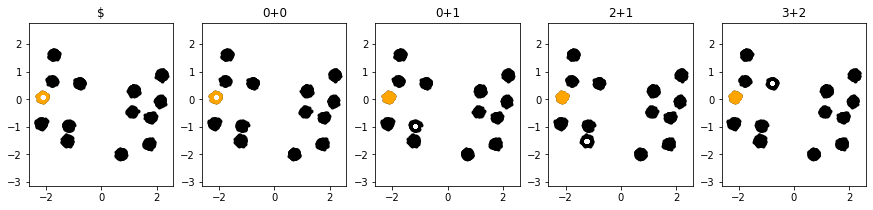}
   \includegraphics[width=0.8\linewidth]{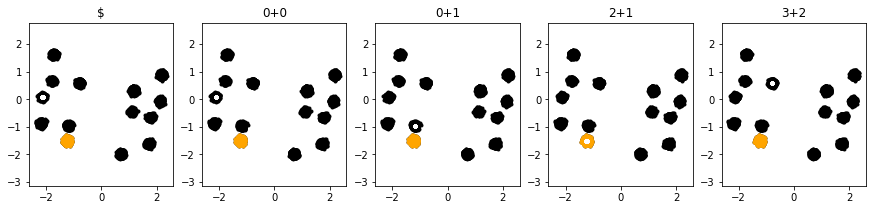} 
   \includegraphics[width=0.8\linewidth]{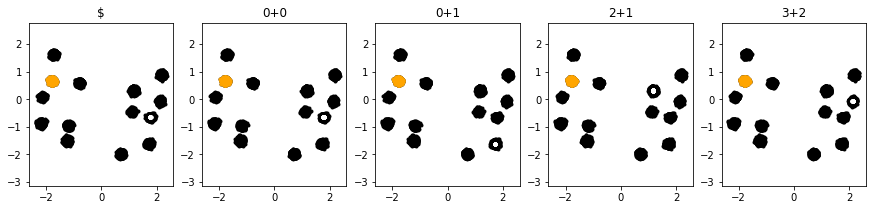}
   \includegraphics[width=0.8\linewidth]{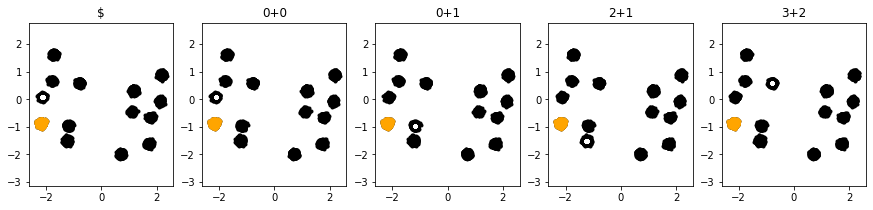}
\end{center}
   \caption{State transitions for the network trained with the base $4$ addition data. The orange dots represent the starting points, the white dots represent the destinations. Each row represents the transitions from one of the first $7$ clusters/states. Each column represents the transitions with an input symbol. Only $5$ different symbols are shown.}
\label{fig::suma_4_all_transitions_b}
\end{figure}

\end{document}